# DE/RM-MEDA: A New Hybrid Multi-Objective Generator

ABEL SAÂ J. R MALANO[1], GUANJUN DU[1], GUOXIANG TONG[1], and NAIXUE XIONG[2, 3], *Senior*, *Member*, *IEEE*

*Abstract*—Under the condition of Karush-Kuhn-Tucker, the Pareto Set (PS) in the decision area of an m-objective optimization problem is a piecewise continuous (*m*-1)-D manifold. For illustrate the degree of convergence of the population, we employed the ratio of the sum of the first (*m*-1) largest eigenvalue of the population's covariance matrix of the sum of all eigenvalue. Based on this property, this paper proposes a new algorithm, called DE/RM-MEDA, which mix differential evolutionary (DE) and the estimation of distribution algorithm (EDA) to generate and adaptively adjusts the number of new solutions by the ratio. The proposed algorithm is experimented on nine tec09 problems. The comparison results between DE/RM-MEDA and the others algorithms, called NSGA-II-DE and RM-MEDA, show that the proposed algorithm perform better in terms of convergence and diversity metric.

## I. Introduction

Multiobjective optimization problems (MOPs) takes place in many engineering field. Over the past three decades, the multi-objective optimizations problems (MOPs) have been presented, but because of their poor hypothesis and global search ability, multi-objective evolutionary algorithms (MOEAs) have been attracting more attention. Various number of MOEAs have been proposed, such as MOGA [1], NSGA [2], NSGA-II [3], PESA [4], PESA-II [5], PAES [6], MOEA/D [7], RM-MEDA [8], Multi-objective evolutionary algorithms produces excellent performance in different proposed field [9]-[11].

In the area of DE, multi-operator hybrid strategies have been deeply concerned. Different operator are used to generate different individuals in each generation, which are the crossover and mutation operators. It has more attention among multi-objective optimization [12]. It has the simply implementation and can solve problems efficiently. Because of its powerful convergence ability and robustness, the differential evolution [13] is suitable for solving m-objective optimization problems in complex environment. Such as NSGA-II-DE [25]. For the control parameters, F (scaling factor) and CR (crossover rate) [14] are set [15]. The distribution estimation algorithm (EDA) [16] establish a probability model and explicitly extracts global information from parent solutions. The new solutions are sampled from [16]-[18]. The EDAs have the ability to deal with hard problems, which are linkage or dependencies among decision variables [22]-[24]. When an inappropriate probabilistic model is selected, EDA become high computational cost and low efficiency. DE an EDA have their own advantages and characteristics do deal with complex MOPs. The RM-MEDA [8] outperform well in various field like a probabilistic model based MOEA. Under the Karush-Kuhn-Tucker (KKT) [19], [20] condition, the Pareto Set of a continuous m-objective optimization problem is a (*m*-1)-D manifold in the decision field [21]. Based on these characteristics, a parameter is introduced, which is the ratio of the sum of the first (*m*-1) largest eigenvalue of the population's covariance matrix to the sum of the all eigenvalue, to indicate the degree of convergence of the population. This paper proposes a new hybrid algorithm called DE/RM-MEDA, which combine the DE and EDA operators for generating some new solution, and adaptively adjust the number of solutions obtained by different operators.

- In DE/RM-MEDA, the information of individual location and the global population distribution information are both used to generate trial solutions by introducing DE and EDA. While RM-MEDA only based on EDA.
- A parameter is introduced in DE/RM-MEDA to illustrate the convergence degree of the population, adjust the number of new solution generated by these two methods.

The remainder of this paper is organized as follow. Section II introduce the multi-objective problems. Section III briefly describes the two other algorithms used in comparison. Section IV presents our proposed algorithm and its specific implementation. In section V, the experimental studies are given to demonstrate the efficiency of the proposed algorithm. Finally, the paper is conclude in section VI.

## II. Problem Definition

In this paper, we consider the following multi-objective problem (MOP):

$$Minimize\ F(x) = \big(f_1(x), \dots, f_n(x)\big)$$
$$subject\ to\ x \in \Omega \tag{1}$$

[1]Shanghai Key Laboratory of Modern Optical System, School of Optical-Electrical and Computer Engineering, University of Shanghai for Science and Technology, Shanghai 200093, China.

[2]College of Intelligence and Computing, Tianjin University, Tianjin 300350, China.
[3]Department of Mathematics and Computer Science, Northeastern State University, Tahlequah, OK 74464, USA.

Where $\prod_{i=1}^{n}[a_i, b_i]$ is the decision space of the parameter $x$ and $x = (x_1, \ldots, x_n)^T \in \Omega$ is a decision variable vector. $F: \Omega \rightarrow R^m$ consist of multi-objective function $f_i(x)$, $i = 1, \ldots, m$, and $R^m$ gives the objective field.

Usually, since the objectives in (1) contradict each other, nil point in $\Omega$ can minimize all the objectives simultaneously. The Pareto-optimal solutions, called Pareto set (PS), are the optimal trade off solutions among different objectives and its mapping to the objective field is the Pareto front (PF).

A vector $u = (u_1, \ldots, u_m)$ s said to dominate another vector $v = (v_1, \ldots, v_m)$ (given by $u \prec v$) if and only if $u_i \leq v_i \,\forall i \in i, \ldots, m$ and $u \neq v$. A point $x^* \in \Omega$ is the Pareto-optimal or non-dominated solution if there is no point such as $x \prec x^*$. That is, any improvement in the Pareto-optimal solution, at least one objective will certainly lead to a deterioration in one other objectives.

## III. THE ALGORITHMS IN COMPARISON

### A. THE RM-MEDA ALGORITHM

We can see from the condition of KKT that under mild conditions, for continuous multi-objective optimization problem, the Pareto set is continuous $(m\text{-}1)$-D piecewise manifold. The population in the decision area will be gradually scattered round the truly PS as the search continues. Based to this idea, in each generation, RM-MEDA [8] at the first time partitions the population into K disjoint clusters to approximate the PS by the $(m\text{-}1)$-D Local PCA algorithm instead of K-means clustering method in which a cluster centroid is a point. Among each clusters, RM-MEDA builds a probabilistic model by extracting the distribution information of parent solutions for the generation of new solutions instead of employing the crossover and mutation operators. In addition, the number of new solutions generated in each cluster is proportional to the volume of the cluster.

a) *Modeling* As the algorithm continues to iterate, the resulting population will increasingly approach the real Pareto optimal solution set, and be distributed evenly around the PS as shown in Fig. 1. In order to describe the distribution of data points in the population as accurately as possible, RM-MEDA establishes a model, and assumes that the points in the population are vectors independent observations of the vector $\zeta$, and the center is $\zeta$. In addition, $\zeta$ is an $m-1$-dimensional continuous manifold in n-dimensional space, and $\varepsilon$ is an n-dimensional noise vector with zero as mean. For simplicity, suppose K segments of continuous manifolds, as shown in fig. 2 approximate that $\zeta$.

$$\zeta_0^0 = \zeta + \varepsilon$$

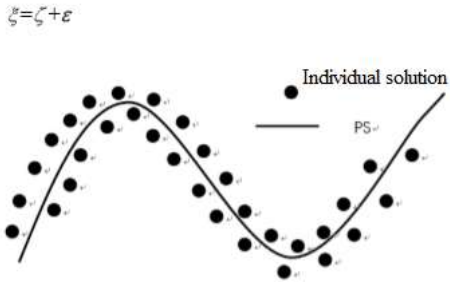

Fig. 1. Schematic diagram of individual solutions distributed around PS

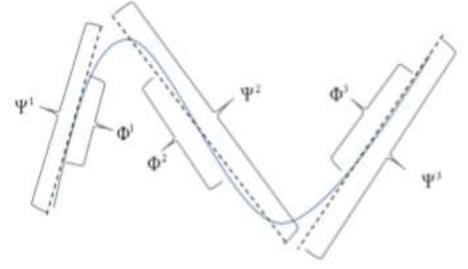

Fig. 2. Two-objective optimization problem and segmentation manifold diagram

In the modeling, the local principal component analysis method first used to divide the population into K clusters, namely $S^1, \ldots, S^k$. The principal component $\Phi^i$ of each cluster is extracted and expanded to $\Psi^i$ to represent the manifold of this cluster, and can provide a better approximation.

b) *Sampling* N new solutions need to be generated during sampling, and the number of new solutions generated by each cluster is proportional to the volume of this manifold $vol(\Psi^i)$. The formula is as follows:

$$Prob(A^j) = \frac{vol(\Psi^j)}{\sum_{i=1}^{k} vol(\Psi^i)} \qquad (2)$$

Use $A^j$ to represent the new individual in the event from the model $\Psi^j$. $vol(\Psi^i)$ indicates the space size of the ith cluster of the extraction manifold. The two-objective problem indicates the length of the line segment. The three-target problem indicates the area, and so on.

The selection use a fast non-dominated sort to select new individuals.

1) The Differential Algorithm (DE), like other evolutionary algorithms, is based on crossover, mutation, selection, etc., and the individuals are iteratively retains with strong environmental adaptability. The basic idea is to recombine the intermediate population with the differences of the current population individuals, and then use the offspring individuals to compete with the parent individuals to obtain a new generation of populations, which has strong global convergence ability and robustness. The most novel feature of the DE algorithm is its mutation operation. After selecting an individual, the difference between the other two individual weights is added to the individual to complete the variation. In the initial stage of the algorithm iteration, the individual differences of the population are large. Such mutation operation will make the algorithm have strong global search ability. By the end of iteration, the population tends to converge, and reduce the differences between individuals, making the algorithm have strong local search ability. It effectively avoids getting into local optimum while also allowing the algorithm to converge faster to the true optimal solution. The specific process is shown in Fig.3.

2) The distribution estimation algorithm is an emerging stochastic optimization algorithm in the field of evolutionary computation, which is obviously different from DE or EDA. It organically combines evolutionary computing and statistical learning. The main idea is to establish a probability model for the distribution of individuals in the solution space by means of statistical methods, and then sample the model to generate a new individual. Different from the traditional evolutionary algorithm, the algorithm does not use crossover and mutation operations. The specific process is shown in Fig.4.

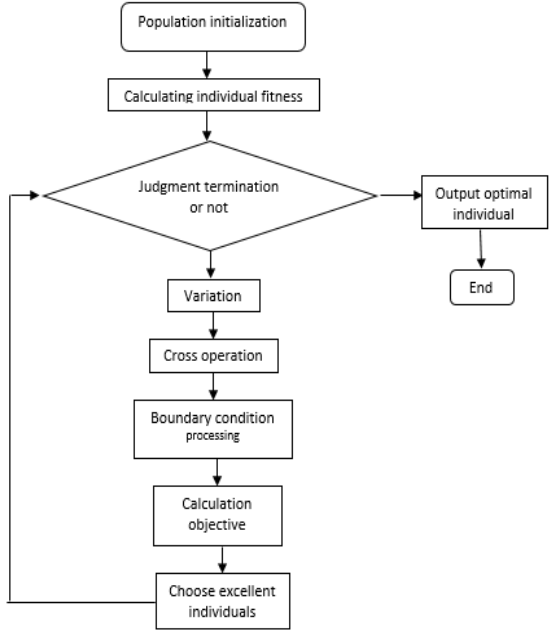

Fig. 3. DE algorithm flow

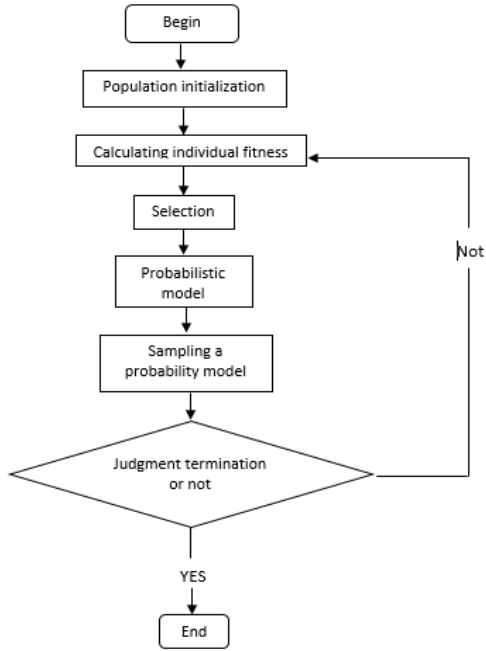

Fig. 4. EDA algorithm flow

## B. NSGA-II

Here, the NSGA-II [3] introduces the evolutionary multi-objective optimization algorithm framework based on Pareto dominance. The main components of NSGA-II include fast non-dominated sorting, congestion calculation, and elite retention strategy.

1) Fast non-dominated sort, the main idea of fast non-dominated sorting is to set an empty $S_p$, and $n_p$ with zero data for each individual p in the population P. Where $S_p$ represents the other individuals in the population dominated by the individual p. $n_p$ represents the number of individuals in the population that dominate the individual p. For each individual q in the population P except the individual p, if p dominates q, put q into the set $S_p$, and if q dominates p, increase the value of $n_p$ by one. When $n_p$ is zero, it means that the individual p is not dominated by any individual and it is a non-dominated solution. Place all individuals with $n_p$ of zero on the first level $F_1$. At this point, we obtain the first layer then continue to stratify other individuals. For each individual p in $F_1$, the $n_p$ of each individual q in the individual set $S_p$ of the individual p is reduced by 1, and if $n_p$ becomes zero at this time, the individual q is divided into the second layer. In addition, put the individual q into the second level set Q. After the individuals in $F_1$ have completed the above operations, the individuals in the set Q are placed in the second layer, and the set Q is then cleared. Repeat the above to get the next level.

2) The calculation of the crowded distance is meaningful in the calculation of the mutual dissatisfaction of the same layer, and the calculation of the crowded distance by the individuals is meaningless in different layers. The crowded distance calculation is a great significance for maintaining the diversity of the solutions when solutions are in selection. The main idea of the crowded distance is that all individual distances of the layer being sorted are set to zero. The individual sets are sorted according to the size of the first objective function value, then the first and the last ranked are set to the infinity distance. The crowded distance value of the individuals in the middle is calculated by finding the previous and next digits of the individual x when sorted according to the objective function value, and calculating the sum of the unit distances between the two digits is the crowded distance value of the individual. Repeat the above process for the other m-1 objective functions, adding the results of the individual x calculations as the final crowded distance of the individual x.

3) Elite retention strategy, the main purpose of the elite retention strategy is to speed up the convergence and reduce the amount of unnecessary calculations. The specific method is to combine the parent population $P_t$ of size N and the newly generated progeny population $Q_t$ as $R_t$. Use the fast non-dominated sorting method to layer $R_t$, starting from the first layer, adding the individual accumulation of each layer, denoted as $F_s$, until $F_s > N$. Put all individuals before the last layer of the calculation into $P_{T+1}$. Remember that the last layer is F1. The crowded distance is sorted for F1, and the calculation of the crowded distance is shown in Fig.6. Individuals with large crowded distances are added to $p_{t+1}$ in turn until the number of individuals in $p_{t+1}$ is N.

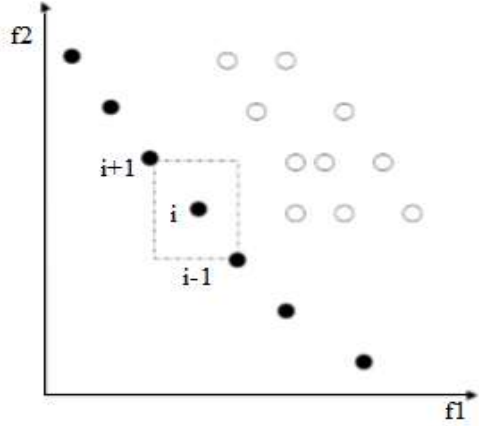

Fig. 5. Schematic diagram of the crowded distance of the individual in the two-dimensional target

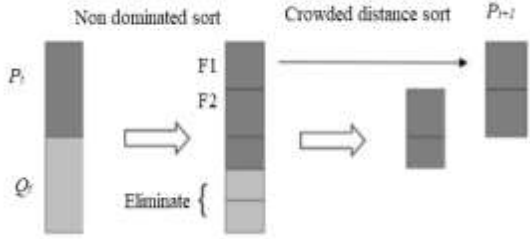

Fig. 6. Execution process of Elite retention strategy

## IV. OUR PROPOSED ALGORITHM

The proposed algorithm in this paper improves the generation stage of RM-MEDA generation, then adds DE operator to generate new individuals, and introduces a parameter to adjust the number of new solutions generated by two operators. In DE, the most novel feature of the algorithm is its mutation operation. After selecting an individual, the difference between the other two individual weights is added to the individual to complete the variation. In the initial stage of the algorithm iteration, the individual differences of the population are large. Such mutation operation will make the algorithm have strong global search ability. By the end of iteration, the population tends to converge, and the differences between individuals are reduced, making the algorithm have strong local search ability. It effectively avoids getting into local optimum while also allowing the algorithm to converge quickly to the true optimal solution.

The algorithm implementation include initialization, intra-cluster clustering, generation of children, selection, etc. are as follows:

| DE/RM-MEDA Framework |
|---|
| **Phase 1: Initialization:** Generate an initial population P and evaluate them |
| **Phase 2: Partition:** Partition P into K clusters by the local PCA algorithm. |
| **Phase 3: Reproduction:** The new solution generated by DE operator and the probability model act as set Q. And their function values of these solutions are computed in Q |
| **Phase 4: Selection:** Select N solution from P ∪ Q as new P based on the NDS-selection of RMMEDA. |
| **Phase 5: Termination:** If the stopping condition is satisfied, export P and their function values as the output of the algorithm and then stop. Otherwise go to **Phase 2** |

### 1) Initialization

Generate an initial population in the data feasible domain using real coding and Latin hypercube sampling $p = \{x_1, x_2, ..., x_N\}$, and calculate is corresponding objective function value.

### 2) Internal clustering of populations

Since the cluster center is a manifold rather than a point, the population is divided into clusters using a local principal component analysis algorithm, which is $\{s^1, s^2, ..., s^K\}$.

For each cluster $S^i$, $i = 1, 2, ..., K$. Calculate the mean and covariance matrix using the following formula $COV$:

$$\overline{x} = \frac{1}{|S^i|} \sum_{x \in S^i} x \tag{3}$$

$$\text{cov} = \frac{1}{|S^i - 1|} \sum_{x \in S^i} (x - \overline{x})(x - \overline{x})^T \tag{4}$$

Will $U^i$ recorded as the $ith$ first feature vector corresponding to large eigenvalue, also known as the covariance matrix of the $ith$ principal component. Then, let $L^{m-1}$ recorded as cluster $S^i$ be the space formed by the points $(m-1)$ affine of the dimension. Its mathematical expression is:

$$\left\{ x \in R^n \middle| x = \overline{x} + \sum_{i=1}^{m-1} \theta_i U^i, \theta_i \in R, i = 1, ..., m-1 \right\} \tag{5}$$

Moreover, the point $x$ to its affine $L^{m-1}$ Euclidean distance $dist(x, L^{m-1})$.

The algorithm framework of principal component analysis (PCA) is as follows:

**Step1.** Initialization includes a random point $L^{m-1}, i = 1, ..., K$

**Step2.**

**a.** The individuals in the population $P$ are divide into clusters according to the following rules $\left\{s^1, s^2, ..., s^K\right\}$.

$$S^j = \left\{x \mid x \in P, and \, dist\left(x, L_i^{m-1}\right) \leq dist\left(x, L_k^{m-1}\right), k \neq j\right\}$$

**b.** Based on the income $S^j$ update $L_i^{m-1}, i = 1, ..., K$.

**c.** If the clustering result changes then go to **Step 2.a** otherwise, stop iterating.

*3) Reproduction*

In each cluster, the DE operator and the Gaussian probability model generate new solutions. Two parameters are introduced $\alpha$ and $\beta$ to adjust the number of new individuals to generate new solutions where $\alpha$ and $\beta$ are both real numbers between [0,1]. The new solution numbers generated by the DE operator and the probability model are respectively $k_1$ and $k_2$. The basic idea of the algorithm is to limit the number of new solutions generated by the DE operator to a certain range. $(\alpha + \beta * p)$ represent the proportion of the new solution generated by the probability model. For example, when $\alpha$ and $\beta$ are 0.1 and 0.7, the maximum and minimum values of new solutions generated by the probability model are 10% and 80%, respectively.

The probabilistic model sampling algorithm flow is as follows:
**Step1.** Create a probability model using the following formula:

$$\psi^j = \left\{x \in R^n \mid x = \overline{x}^j + \sum_{i=1}^{m-1} a_i U_i^j, a_i^j - 0.25\left(b_i^j - a_i^j\right) \right.$$
$$\left. \leq a \leq b_i^j + 0.25\left(b_i^j - a_i^j\right), i = 1, ..., m-1\right\} \quad (6)$$

Among them,

$$a_i^j = \min_{x \in S^j}(x - \overline{x}^j)^T U_i^j,$$
$$b_i^j = \max_{x \in S^j}(x - \overline{x}^j)^T U_i^j \quad (7)$$

**Step2.** Record $\lambda_i^j$ as the largest eigenvalue of the cluster. Set

$$\sigma_j = \frac{1}{n - m + 1} \sum_{i=m}^{n} \lambda_i^j \quad (8)$$

**Step3.** Generate a point $\overline{x}$ from $\psi^j$ and a noise vector $\overline{\varepsilon}$ from $N\left(0, \sigma_j I\right)$, where $I$ is the $n \times n$ identity matrix.

**Step4.** Return the new solution $\overline{x} + \overline{\varepsilon}$.

The reproduction algorithm framework is as follows:
**Step1.** Calculate the sum of the first $(m-1)$ eigenvalues of the population covariance matrix and the sum of all eigenvalues in each cluster.

$$\sigma_{m-1} = \sum_{i=1}^{m-1} \lambda_i^j, \sigma_n = \sum_{i=1}^{n} \lambda_i^j, p = \frac{\sigma_{m-1}}{\sigma_n} \quad (9)$$

**Step2.** Let

$$k_1 = N \times \frac{vol(\psi^j)}{\sum_{i=1}^{k} vol(\psi^j)} \times (\alpha + \beta \times p) \quad (10)$$

$$k_2 = N \times \frac{vol(\psi^j)}{\sum_{i=1}^{k} vol(\psi^j)} - k_1 \quad (11)$$

Where $vol(\psi^j)$ is the volume of the $jth$ $(m-1)-$ dimensional manifold, and

$$N \times \frac{vol(\psi^j)}{\sum_{i=1}^{k} vol(\psi^i)} \quad (12)$$

Indicates the number of new solutions generated by the $jth$ cluster.

**Step3.** Randomly select $k_2$ solution in the $jth$ cluster to generate new individuals by using the DE generation operator.

**Step4.** Generate $k_1$ new solutions by using the probability model.

The differential algorithm (DE) uses the cross and mutation operations to generate the children. Different versions of the differential algorithm were proposed.

The DE algorithm mechanism is as follows:
**Step1.** Randomly select the parent individual from the cluster to satisfy the following conditions: $x^{r_1} \neq x^{r_2} \neq x^{r_3} \neq x^{best}$ where $x^{best}$ indicates that, in the current population there are not any Individuals dominated by other individuals.

**Step2.** Generate intermediate individuals $v$ by the following method:

$$v_k = \begin{cases} x_k^{r_1} + F(x_k^{best} - x_k^{r_1}) + F(x_k^{r_2} - x_k^{r_3}) \ if \ r < CR \\ x_k^{r_1} \quad otherwise \end{cases} \quad (13)$$

Where $r$ is the random number between [0,1].

**Step3.** Perform the following mutation on $v$ to generate a new solution $x$.

$$x_k = \begin{cases} v_k + \delta \times (u_k - l_k) \ \ if \ r < P_m \\ v_k \quad otherwise \end{cases} \quad (14)$$

With

$$\delta = \begin{cases} (2 \times r)^{\frac{1}{\eta+1}} - 1 \ \ if \ r < 0.5 \\ 1 - (2 - 2 \times r)^{\frac{1}{\eta+1}} \quad otherwise \end{cases} \quad (15)$$

Where $\mu_k$ and $l_k$ are the upper and lower bounds of the $kth$ decision space.

### 4) Selection

Based on the NDS-selection of RMMEDA, $N$ individuals from $P \cup Q$ are selected as new population $P$ by quick sort.

### 5) Termination:

When the stopping condition is satisfied, the population P and their function values are exported as the output of the algorithm and then stop, otherwise go to internal clustering of population.

## V. PERFORMANCE ANALYSIS

In the experimental environment, the algorithm is modeled and simulated by MATLAB, in order to verify that the proposed algorithm outperforms better than other multi-objective optimization algorithms and the solution to clustering problems, the performance of the algorithm was verified by using the tec09 test set.

### PERFORMANCE ANALYSIS BASED ON TEC09 TEST SET

#### 1) Figure test functions and parameters settings

In this section, the proposed algorithm is applied to nine test functions, F1 to F9 with complex Pareto set proposed in [15]. As comparison algorithms, we selected RM-MEDA [8] and NSGA-II-DE [25]. Where F6 has three objective functions and the other has two. The Front Pareto (PF) of F6 and F9 is non-convex, and the PF of other functions is convex. F7 and F8 have multiple local optimum PFs and are multi-peak problems. The independent variable dimension of F1-F5 and F9 is 30, and the other functions are 10. Each algorithm runs for 500 generations. The population size of the function F6 is 600, and the other functions are 300.

We chose IGD (Inverted Generational Distance) as the performance evaluation indicator. Its mathematical expression is:

$$\begin{cases} IGD(P, P^*) = \dfrac{\sum v \in P^*\, d(v, p)}{\left| p^* \right|} \\ d(x, P^*) = \min_{y \in P^*} \| x - y \| \end{cases} \quad (16)$$

Where $P^*$ indicates the true Pareto Front, a set of solutions obtained by uniformly sampling the PF. $P$ represents the approximation solution set obtained by the algorithm. The larger $|P^*|$ the more representative of true Pareto Front, $IGD(P, P^*)$ indicator can simultaneously measure the diversity and convergence of the population $P$. The $P^*$ for the two-objective problem in this paper usually consists of 500 points, while the three-target problem consists of 1000 points. All algorithms will run 20 times independently to ensure the fairness of the algorithm. In DE and polynomial mutation, the cross probability $CR = 0.9$, the variation rate of the scaling factor $F = 0.5$, $\eta = 20$ and $p_m = 1/n$. In the algorithm DE/RM-MEDA, the parameters introduced to adjust the number of new solution α and β are respectively 0.3 and 0.6 and the number of clusters K= 5.

#### 2) Figure analysis of experimental parameters

The main control parameters of the DE/RM-MEDA algorithm include the number of clusters, the adjustment parameters $\alpha$ and $\beta$, and the decision space dimension.

First, test the effect of the number of clusters on the performance of the algorithm, the experiment selected F6 as a test function. The experiment setting of the number of clusters K = 3, 5, 7, 9, 11, 13, 15. Fig. 7 shows the IGD-Metric mean of the PS obtained by DE/RM-MEDA in the case of different objective function, calculation times and different cluster numbers. The results show that the algorithm DE/RM-MEDA obtains similar results in the number of clusters in different populations. The number of clusters within the population does not affect this algorithm. In addition, the algorithm after 5000 iterations is able to reduce the IGD-Metric under 0.1.

Second, consider the impact of the search space dimension on the performance of the algorithm. We set the argument dimension d=30, 50,100. The results obtained by the three algorithms on the test functions F1 to F5 and F9 are ranked as shown in Fig. 8 it can be seen from the figure that the performance of the DE/RM-MEDA algorithm is not affected by the variation of the independent variable dimension, and the best results are obtained in the three algorithms.

Finally, consider the effects of different adjustment parameters α and β on the performance of the algorithm. Fig.9. shows the experimental. We set α and β to 0.1-0.7, 0.2-0.6, and 0.3-0.6 from left, to right corresponding to three data for each test function in the graph. In the fig. 9, the blue are set to α = 0.1, β = 0.7, the green are set to α = 0.2, β = 0.6, and the yellow are α = 0.3, β = 0.6. The results show that the algorithm can achieve the best results when α and β are respectively set to 0.3 and 0.6.

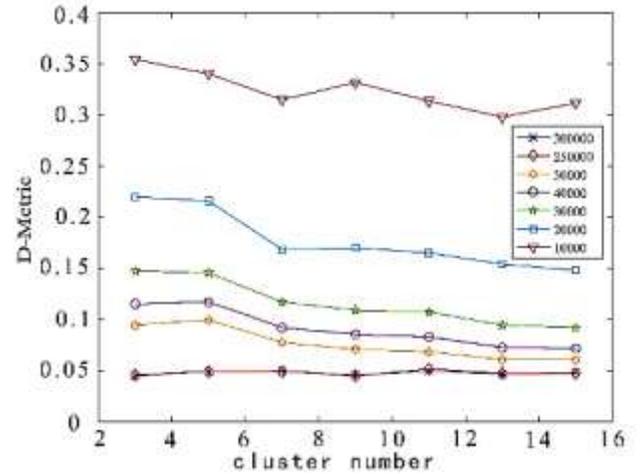

Fig. 7. The mean D-Metric value produced by DE/RM-MEDA for F6 include different numbers of cluster.

|  | NSGA-II-DE | RM-MEDA | DE/RM-MEDA |
|---|---|---|---|
| F1 | 0.0044(0.0000) | 0.0015(0.0000) | 0.0015(0.0000) |
| F2 | 0.0349(0.0066) | 0.0514(0.0048) | 0.0482(0.0056) |
| F3 | 0.0296(0.0030) | 0.0140(0.0010) | 0.0142(0.0013) |
| F4 | 0.0288(0.0021) | 0.0211(0.0039) | 0.0183(0.0023) |
| F5 | 0.0288(0.0031) | 0.0128(0.0011) | 0.0107(0.0011) |
| F6 | 0.0680(0.0072) | 0.0483(0.0065) | 0.0446(0.0066) |
| F7 | 0.1171(0.0716) | 0.0555(0.0404) | 0.0401(0.0414) |
| F8 | 0.1981(0.0494) | 0.0834(0.0244) | 0.0815(0.0310) |
| F9 | 0.0395(0.0061) | 0.0480(0.0032) | 0.0511(0.0038) |

TABLE II
THE MEAN AND STD IGD VALUES GENERATED BY DE/RM-MEDA
ALGORITHM ON SIX TEST INSTANCES WITH DIFFERENT VARIABLE
DIMENSIONS.

|  | 100 | 200 | 300 | 400 | 500 |
|---|---|---|---|---|---|
| F1 | 0.0019 | 0.0015 | 0.0015 | 0.0015 | 0.0015 |
|  | (0) | (0) | (0) | (0) | (0) |
| F4 | 0.0949 | 0.0501 | 0.0340 | 0.0264 | 0.0183 |
|  | (0.0093) | (0.0060) | (0.0044) | (0.0031) | (0.0023) |
| F5 | 0.0656 | 0.0311 | 0.0214 | 0.0173 | 0.0107 |
|  | (0.0047) | (0.0025) | (0.0019) | (0.0012) | (0.0011) |
| F6 | 0.1571 | 0.1068 | 0.0628 | 0.0503 | 0.0446 |
|  | (0.0085) | (0.0082) | (0.0071) | (0.0068) | (0.0066) |
| F7 | 0.4849 | 0.2148 | 0.1003 | 0.0604 | 0.0401 |
|  | (0.1599) | (0.1090) | (0.0756) | (0.0519) | (0.0414) |
| F8 | 0.4280 | 0.2987 | 0.1812 | 0.1067 | 0.0815 |
|  | (0.0844) | (0.0487) | (0.0705) | (0.0512) | (0.0310) |

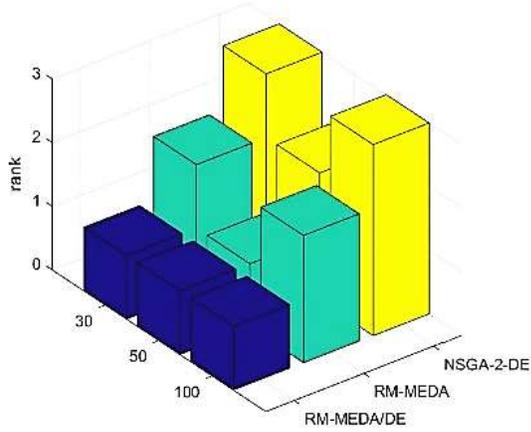

Fig. 8. The ranks results generated by the three algorithms on 6 test instances with the d=30, 50,100

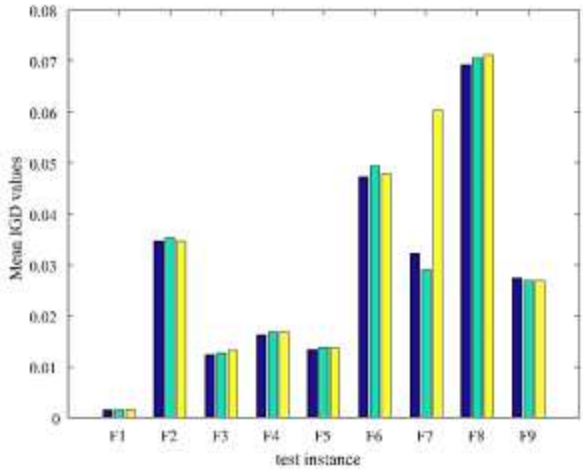

Fig. 9. The result of different IGD values generated by DE/RM-MEDA on the 9 test instances with different $\alpha$ and $\beta$ parameter.

*1) Algorithm comparison experiment*

Table. I shows the results of the statistical IGD mean value obtained by DE/RM-MEDA, RM-MEDA, and NSGA-II-DE (the results of NSGA-II-DE comes from the literature [8]). We can see from Table. I that DE/RM-MEDA achieves better results on F4, F5, F6, F7, and F8. NSGA-II-DE achieved the best results on F2 and F9. RM-MEDA achieved the best results on F3 and have the same performance with DE/RM-MEDA on F1, maybe because F1 is not complicated so RM-MEDA can perform well. The data results show that the simple DE and the simple probability model operators cannot achieve good results, but when they are combined together to form a hybrid algorithm, we can obtain better convergence and diversity. Combining individual information with population distribution information can enhance the performance of the algorithm especially for test functions F7 and F8, DE/RM-MEDA handles multi-peak problems with better performance. When we compare the IGD values obtained by the three algorithms on the function F6, we can conclude that the operator mixing different information is more advantageous for processing the high-dimensional multi-objective function.

Table. II gives the results of the statistical IGD indicators of DE/RM-MEDA on the six test functions. The results show that the IGD metric decreases with the increase of algebra, and the change is most obvious before the first 300 generations. That means the DE/RM-MEDA algorithm can efficiently find a stable solution set. After 500-generation iterations, all problems reached a small IGD mean. Except for F7 and F8, other problems can be obtained with a smaller IGD mean square error. Maybe because F7 and F8 are multi-peak problems, and the existence of more local optimums makes the algorithm difficult. However, compared to F8, the algorithm is able to get a smaller IGD mean on F7.

## VI. CONCLUSION

It has not be well studied how to utilize the individual and population distribution information efficiently to generate new trial solutions based on the framework of non-dominated sorting. Meanwhile, the ratio of the sum of the first (m-1) largest eigenvalue of the population's covariance matrix to the sum of the whole eigenvalue has not been employed to illustrate the degree of convergence of the population. A hybrid algorithm named DE/RM-MEDA is proposed in this paper for multi-objective problems with complicated PS shapes. In our approach, the Gaussian probabilistic model based operator and the DE operator are employed to generate new solutions collaboratively. At each generation, the number of new solutions generated by the two operators is adjusted by the ratio. The bigger the ratio is, the more new solutions will generated by Gaussian probabilistic model. The new algorithm combines individual and population information efficiently to balance the exploitation and the exploration search abilities. Experimental studies have shown that overall, the new algorithm proposed in this paper performs better than NSGA-II-DE and RM-MEDA on a set of test instances with complicated PS shapes. The proposed algorithm can approach the real Pareto frontier faster and more effectively than the RM-MEDA using the distribution estimation algorithm alone and the NSGA-II-DE using the differential algorithm alone.

## REFERENCES

[1] C. M. Fonseca and P. J. Fleming, ''Genetic algorithms for multiobjective optimization: Formulation discussion and generalization,'' ICGA, vol. 93, pp. 416–423, Jul. 1993.

[2] N. Srinivas and K. Deb, ''Muiltiobjective optimization using nondominated sorting in genetic algorithms,'' Evol. Comput., vol. 2, no. 3, pp. 221–248, 1994.

[3] K. Deb, A. Pratap, S. Agarwal, and T. Meyarivan, ''A fast and elitist multiobjective genetic algorithm: NSGA-II,'' IEEE Trans. Evol. Comput., vol. 6, no. 2, pp. 182–197, Apr. 2002

[4] D. W. Corne, J. D. Knowles, and M. J. Oates, ''The Pareto envelopebased selection algorithm for multiobjective optimization,'' in Proc. Int. Conf. Parallel Problem Solving Nature. Berlin, Germany: Springer, 2000, pp. 839–848.

[5] D. W. Corne, N. R. Jerram, J. D. Knowles, and M. J. Oates, ''PESA-II: Region-based selection in evolutionary multiobjective optimization,'' in Proc.3rdAnnu.Conf.Genet.Evol.Comput.SanMateo,CA,USA:Morgan Kaufmann, 2001.

[6] J. Knowles and D. Corne, ''Approximating the nondominated front using the Pareto archived evolution strategy,'' Evol. Comput., vol. 8, no. 2, pp. 149–172, Jun. 2000.

[7] Q. Zhang and H. Li, ''MOEA/D: A multiobjective evolutionary algorithm based on decomposition,'' IEEE Trans. Evol. Comput., vol. 11, no. 6, pp. 712–731, Dec. 2007.

[8] Q. Zhang, A. Zhou, and Y. Jin, ''RM-MEDA: A regularity model-based multiobjective estimation of distribution algorithm,'' IEEE Trans. Evol. Comput., vol. 12, no. 1, pp. 41–63, Feb. 2008.

[9] E. C. Hinojosa and H. A. Camargo, ''Multi-objective evolutionary algorithm for tuning the Type-2 inference engine on classification task,'' Soft Comput., vol. 22, no. 15, pp. 5021–5031, 2018.

[10] S. Acharya, S. Saha, and P. Sahoo, ''Bi-clustering of microarray data using a symmetry-based multi-objective optimization framework,'' in Soft Computing, May 2018, pp. 1–22.

[11] A. S. Pillai, K. Singh, V. Saravanan, A. Anpalagan, I. Woungang, and L. Barolli, ''A genetic algorithm-based method for optimizing the energy consumption and performance of multiprocessor systems,'' Soft Comput., vol. 22, no. 10, pp. 3271–3285, 2017.

[12] S. Wang, Y. Li, H. Yang, and H. Liu, ''Self-adaptive differential evolution algorithmwithimprovedmutationstrategy,''SoftComput.,vol.22,no.10, pp. 3433–3447, 2018.

[13] S.Das,S.S.Mullick,andP.N.Suganthan,''Recentadvancesindifferential evolution—An updated survey,'' Swarm Evol. Comput., vol. 27, pp. 1–30, Apr. 2016.

[14] Q. Wei and X. Qiu, ''Dynamic Differential Evolution algorithm with composite strategies and parameter values self-adaption,'' inProc. 7th Int. Conf. IEEE Adv. Comput. Intell. (ICACI), Mar. 2015, pp. 271–274.

[15] N. J. Singh, J. S. Dhillon, and D. P. Kothari, ''Multi-objective thermal power load dispatch using chaotic differential evolutionary algorithm and Powell's method,'' Soft Comput., vol. 22, no. 7, pp. 2159–2174, 2018.

[16] H. Mühlenbein and G. I. Paass, ''From recombination of genes to the estimation of distributions I. Binary parameters,'' in Proc. Int. Conf. Parallel Problem Solving Nature. Berlin, Germany: Springer, 1996, pp. 178–187.

[17] G. R. Harik, F. G. Lobo, and D. E. Goldberg, ''The compact genetic algorithm,''IEEETrans.Evol.Comput.,vol.3,no.4,pp.287–297,Nov.1999.

[18] S.Baluja,''Population-basedincrementallearning.Amethodforintegrating genetic search based function optimization and competitive learning,'' Dept. Comput. Sci., Carnegie-Mellon Univ., Pittsburgh, PA, USA, 1994.

[19] K. Miettinen, Nonlinear Multiobjective Optimization, vol. 12. Springer, 2012.

[20] O. Schütze, S. Mostaghim, and M. Dellnitz, ''Covering Pareto sets by multilevel evolutionary subdivision techniques,'' in Proc. Int. Conf. Evol. Multi-Criterion Optim. Berlin, Germany: Springer, 2003, pp. 118–132.

[21] Hillermeier C.2001 Nonlinear multiobjective optimization: a generalized homotopy approach Springer Science & Business Media, 2001.

[22] S. Baluja and S. Davies, ''Using optimal dependency-trees for combinatorial optimization: Learning the structure of the search space,'' Dept. Comput. Sci., Carnegie-Mellon Univ, Pittsburgh, PA, USA, 1997.

[23] G. Harik, ''Linkage learning via probabilistic modeling in the ECGA,'' Urbana, vol. 51, no. 61, p. 801, 1999.

[24] M. Pelikan, D. E. Goldberg, and E. Cantú-Paz, ''Linkage problem, distribution estimation, and Bayesian networks,'' Evol. Comput., vol. 8, no. 3, pp. 311–340, 2000.

[25] H. Li and Q. Zhang, ''Multiobjective optimization problems with complicated Pareto sets, MOEA/D and NSGA-II,'' IEEE Trans. Evol. Comput., vol. 13, no. 2, pp. 284–302, Apr. 2009.